# Limited-Angle CBCT Reconstruction via Geometry-Integrated Cycle-domain Denoising Diffusion Probabilistic Models

Yuan Gao[†,1], Shaoyan Pan[†,1], Mingzhe Hu[†,1], Huiqiao Xie[1], Jill Remick[1], Chih-Wei Chang[1],

Justin Roper[1], Zhen Tian[2*] and Xiaofeng Yang[1*]

[1]Department of Radiation Oncology and Winship Cancer Institute, Emory University, Atlanta, GA
[2]Department of Radiation & Cellular Oncology, University of Chicago, Chicago, IL

**Running title:** Limited-Angel CBCT Reconstruction via DDPM

*Corresponding to: ztian@bsd.uchicago.edu and xiaofeng.yang@emory.edu

**Manuscript Type:** Original Research

†These authors contributed equally to this work.

# Abstract

**Purpose:** Cone-beam computed tomography (CBCT) has been widely adopted in clinical radiation therapy for its capability to enable image-guided treatment delivery. Its applications—such as improving setup accuracy, facilitating anatomy-of-the-day–based adaptive planning, and enabling motion management—have significantly enhanced treatment precision and confidence.

However, the relatively slow gantry rotation (60-90 seconds) is one of the main limiting factors, which causes motion artifacts, blurred anatomy in the images, elevated imaging dose, etc. Therefore, a short-arc scan with limited-angle projections to reduce the imaging time is highly desirable. The objective of this work is to develop a clinically feasible method for reconstructing high-quality CBCT volumes from consecutive limited-angle acquisitions, address imaging constraints in time-sensitive or dose-limited clinical settings.

**Methods and Materials:** We propose a limited-angle (LA) Geometry-integrated Cycle-domain (LA-GICD)-framework for LA CBCT reconstruction. The proposed LA-GICD framework includes two denoising diffusion probabilistic models (DDPM) that are connected by analytic cone-beam forward- and back-projectors. First, a Projection-DDPM fills the missing projections. These completed sinogram is then back-projected, and finally an Image-DDPM refines the resulting volume images. This dual-domain, geometry-integrated design exploits complementary priors from both projection and image spaces, delivering quality reconstructions from limited-angle ($\leq 90°$) scan. The proposed network's performance was evaluated against reconstructions from full-angle (360°) scans in terms of mean absolute error (MAE), structural similarity index (SSIM), and peak signal-to-noise ratio (PSNR). Qualitative assessment by four board-certified medical physicists visually examined the images to confirm these quantitative gains, focusing on soft-tissue boundaries and limited-angle artifacts. 78 planning CT images were simulated in common CBCT geometries which are the major dataset used for the training and evaluation of the proposed framework.

**Results:** The method achieved a mean absolute error of 35.5 HU, SSIM of 0.84, and PSNR of 29.8 dB, with visible reductions in limited-angle (streak and shading) artifacts and improved soft-tissue clarity. Performance was consistent across axial, coronal, and sagittal views. LA-GICD embeds geometry-aware dual-domain learning in an analytic forward-/backward- projection operator pair, yielding artifact-free, high-contrast reconstructions from a single 90° scan that reduces acquisition time (and dose) by four-fold without needs for hardware changes/upgrades.

**Conclusions:** LA-GICD enables improved LA CBCT reconstruction with strong data fidelity and high anatomical realism. It possesses the potential to offer a practical solution for short-arc CBCT image reconstruction. The potential application could further improve the usage of CBCT in radiotherapy by providing clinical applicable CBCT images with reduced scanning time and imaging dose to promote more accurate, personalized treatments.

## 1. Introduction

Cone-beam computed tomography (CBCT) has become an essential imaging modality in modern radiation therapy. Mounted on linear accelerator and C-arm gantries, flat-panel CBCT systems provide in-room volumetric imaging that enables sub-millimeter patient setup verification, adaptive replanning based on the anatomy-of-the-day (intraoperative in room brachytherapy), and intra-fraction motion monitoring[1]. For external beam treatments, daily CBCT reduces setup errors and improves target coverage with daily anatomy verification. However, slow CBCT acquisitions, which typically require 60–90 seconds of gantry rotation, introduce several clinically significant limitations. Respiratory and internal organ motion during this period degrade image quality through blurring and ghosting[2-4]. In thoracic regions, lung lesions and vessel walls may appear distorted due to breathing cycles; in abdominal and pelvic regions, peristalsis and bowel gas can lead to inconsistent contrast and motion artifacts. Involuntary patient movements—such as muscle twitches, discomfort-related shifts, or coughing—further contribute to streaking and misregistration, compromising spatial accuracy, which leads to elevated margins during treatment planning. In interventional workflows, full-scan CBCT presents additional challenges. The prolonged acquisition time can interrupt procedural steps such as needle or applicator placement and often necessitate extended sedation or general anesthesia, increasing the risk of anesthetic complications[5]. In addition to time and motion-related challenges, full-arc CBCT is often limited by spatial constraints[6]. In operating rooms, brachytherapy suites, and interventional radiology settings, equipment such as anesthesia machines, sterile drapes, and applicators may obstruct the gantry's path, making full 360° rotation infeasible. As a result, image acquisition is often restricted to partial arcs to avoid collisions between the gantry, couch, and patient. Although alternative acquisition strategies, such as non-circular or flexible trajectories, may reduce collision risk and mitigate metal artifacts, they often require specialized calibration and are not widely implemented. These limitations can disrupt workflow, increase setup variability, and reduce image quality, particularly for mobile patient anatomy or in a crowded procedural environment. To address these challenges, limited-angle CBCT acquisition has gained increasing attention as a potential alternative to full-arc scanning. By restricting the gantry rotation to a single arc segment—typically 90° or less—clinicians can reduce scan time, minimize motion artifacts, and avoid mechanical collisions, while using standard imaging hardware.

However, the incomplete projection data introduces artifacts—manifesting as shading, streaking, and blurring—that degrade image quality. For instance, the Feldkamp-Davis-Kress (FDK)[7] the most widely used analytical algorithm—is accurate and efficient with complete projection data, but under limited-angle acquisition it produces shadow artifacts because of the missing information[8]. To address this challenging problem, a variety of approaches have been developed for limited-angle CBCT reconstruction. These methods are generally categorized into two groups: model-based iterative reconstruction (MBIR), and data-driven deep learning frameworks. However, each approach has distinct limitations: MBIR addresses the data-incompleteness problem but brings new challenges, notably a heavy computational burden and strong dependence on carefully tuned regularization parameters. For example, total-variation–regularized (TV) MBIR can suppress streak artifacts, yet an ill-chosen TV weight often produces staircase effects in homogeneous regions and blurs fine anatomical detail[9,10]. Deep-learning based reconstruction algorithm have recently shown great promise, yet they introduce a new set of limitations—most notably the risk of hallucinated anatomy[11], inadequate enforcement of data consistency[12], and poor generalizability outside the training domain.

Recent deep-learning efforts have substantially advanced limited-angle CBCT reconstruction. Early work treated the task as post-image processing, employing convolutional neural networks (e.g., U-Net) to transform streak-laden FBP or FDK reconstructions into artifact-reduced image[13,14]. However, because the projection data are no longer enforced during inference, these post-processing networks often leave residual streaks and may hallucinate anatomy, motivating the shift toward physics-guided iterative networks. To strengthen data consistency, subsequent work unfolded MBIR into trainable deep networks, mapping each iterative update onto a learnable block that jointly enforces data fidelity and regularization[15-17]. Although this physics-guided design suppresses streaks more effectively than image-domain CNNs, its performance remains highly sensitive to predefined step sizes and regularization weights, depends on large, paired datasets tailored to a particular scanning geometry, and can still leave staircase artifacts or residual streaks when acquisition conditions differ from those seen during training. Most recently, researchers have introduced attention-based and Transformer-hybrid networks to solve this problem. By inserting self-attention blocks—often in a Swin (Shifted Window Vision Transformer) or ViT (Vision Transformer) configuration—these models capture long-range correlations that help restore extended anatomical structures in limited-angle data[18]. Hybrid designs combine convolutional layers for fine textures with transformer attention for global context, achieving clearer organ boundaries and fewer streaks than either post-processing CNNs or unfolded MBIR networks. Even so, their performance still depends on large, anatomy-specific training sets, and the substantial model size can lead to overfitting and domain-shift vulnerability when the acquisition geometry or patient population differs from that used for training. In parallel, diffusion-based generative models have been investigated for limited-angle CT/CBCT[19,20]. These networks learn a stepwise denoising process that can inpaint the missing sinogram wedge or synthesize a full image from noise, producing sharp textures and allowing uncertainty sampling. Yet when a diffusion model is tasked with jumping directly from incomplete projections to a reconstructed CT/CBCT image, it operates with only weak physics-based conditioning, making it easier for hallucinated structures to appear. As a result, there is still no limited-angle CT/CBCT method that both follows the measured data closely and avoids creating false structures. This study aims to fill this gap.

In this study, we propose a Limited-angle Geometry-integrated Cycle-domain (LA-GICD) framework for LA CBCT reconstruction that combines three key components into a unified pipeline. First, a projection-domain denoising diffusion model (Projection-DDPM) learns to complete missing projections in a data-driven manner, without requiring handcrafted priors. Second, a geometry transformation module (GTM) projects the completed sinogram into image space using the known cone-beam scan geometry, thereby ensuring physical consistency. This step defines the "geometry-integrated" aspect of the framework and warrants explicit clarification to emphasize its role in preserving ray fidelity. Third, an image-domain CBCT reconstruction denoising diffusion model (Image-DDPM) further refines the reconstructed CBCT volume by suppressing artifacts and restoring fine anatomical detail. Here, the term "cycle-domain" refers to the projection→image→projection loop enforced via analytic forward and back-projection operators, ensuring that the completed sinograms can be reconstructed into volumes that, when re-projected, faithfully match the measured data. By enforcing cyclic consistency between the projection and image domains, the LA-GICD framework enables more accurate and robust reconstructions. This dual-domain, geometry-aware design improves generalization in real-world limited-angle settings, where conventional methods often fail due to undersampling and lack of physical constraints. The key methodological contributions are summarized as follows.

- LA-GICD. We present a limited-angle CBCT framework capable of reconstructing volumetric images from a single 90° acquisition arc using one unified model.
- GTM. A fixed, analytic cone-beam projector/back-projector injects exact geometric priors at every iteration, consistently aligning projection and image spaces and enhancing robustness across scanners and angle deficits.

- GICD strategy. Two linked diffusion modules—a Projection-DDPM for projection completion and an Image-DDPM for image denoising and refinement—operate in a closed projection–image–projection loop that alternates global inpainting and fine-detail refinement, enforcing consistency with measured data and preventing hallucinated anatomy.

## 2. Materials and Methods

### 2.1 LA-GICD framework

#### 2.1.1. Overview

We propose a novel deep generative framework, named LA-GICD, for high-fidelity CBCT reconstruction from limited-angle projections. The proposed architecture mitigates the ill-posedness of limited-angle reconstruction by integrating data-driven priors with explicit geometric consistency constraints.

The pipeline contains three fully differentiable blocks: Projection-DDPM first converts 135°–225° sinograms into full-view projections through conditional denoising diffusion; the GTM then maps these projections to CBCT volumes by filtered back-projection while preserving scanner geometry; Image-DDPM finally refines the GTM output, removing residual artifacts and sharpening anatomy. A GICD loss enforces bidirectional consistency between projection and image spaces. Both DDPMs share one noise schedule, and GTM lets gradients flow from the final image loss back to projection synthesis. Training uses pixel-wise CBCT reconstruction loss, projection cycle loss, and edge-aware penalties that keep high-frequency anatomy, so the network can fill missing angles yet still respect geometric fidelity. A similar cycle-domain, geometry-integrated diffusion design was first demonstrated in our earlier patient-specific CBCT frameworks, including the Cycle-domain Geometry-integrated DDPM (CG-DDPM)[21] model for single-view CBCT reconstruction and the Patient-specific Physics-integrated DDPM (PC-DDPM) approach for real-time Optical Surface-Derived CBCT (OSD-CBCT)[22] synthesis.

#### 2.1.2 Projection-DDPM

The Projection-DDPM module is de designed to generate a complete set of synthetic projections $\hat{P}_{1:360°}$ from limited-angle projections $\hat{P}_{135°:225°}$. This task is formulated as a conditional generative process within the DDPM framework. The model learns to map from Gaussian noise to high-fidelity, full-view projections conditioned on the limited-angle projections input. In the DDPM network, the forward process gradually perturbs a clean data sample with Gaussian noise through a Markov chain[23]. In the forward process, a clean projection image $P$ is gradually perturbed by Gaussian noise through a Markov chain:

$$q(x_t|x_{t-1}) = \mathcal{N}\left(P_t; \sqrt{1-\beta_t}\, P_{t-1}, \beta_t \mathbf{I}\right) \quad (1)$$

With a predefined noise schedule $[\beta_t]_{t=1}^T$. The marginal distribution at any time step $t$ can be written as:

$$q(P_t|P_0) = \mathcal{N}\left(P_t; \sqrt{\bar{\alpha}_t}\, P, (1-\bar{\alpha}_t)\mathbf{I}\right) \quad (2)$$

where $\alpha_t = 1 - \beta_t$ and $\overline{\alpha_t} = \prod_{s=1}^t \alpha_s$.

The reverse process is parameterized by a neural network $\epsilon_\theta$ to predict the noise added at each step:

$$P_{t-1}^{syn} = \frac{1}{\sqrt{\alpha_t}}\left(P_t^{syn} - \frac{1-\alpha_t}{\sqrt{1-\overline{\alpha_t}}} \cdot \epsilon_\theta(P_t^{syn}, t, P_0^{real})\right) + \sigma_t \cdot z, \quad z \sim \mathcal{N}(0,I) \quad (3)$$

The model is trained to minimize a mean squared error between the predicted noise and true noise:

$$\mathcal{L}_{Proj-DDPM} = \mathbb{E}_{P_0^{syn},\epsilon,t}\left[\left\|\epsilon - \epsilon_\theta\left(P_t^{syn}, p_0^{real}, t\right)\right\|_2^2\right] \quad (4)$$

To ensure anatomical plausibility under the limited-angle constraint, we condition the denoising network $\epsilon_\theta$ on the available real projection data. Specifically, the U-Net–based denoiser receives two inputs at each denoising step: the noisy synthesized projection (full view) $P_t^{syn}$ and the real limited-angle projection $P_0^{real}$. The latter acts as a geometric and anatomical prior, providing essential constraints that guide the synthesis toward physically plausible solutions. This conditioning encourages consistency with the measured data while allowing the network to infer missing information beyond the limited angular coverage. The final output of this module is a clean synthesized projection $P_0^{syn}$, which approximates a full-view projection. These synthesized projections are subsequently passed to the GTM to produce intermediate CBCT volumes for downstream refinement.

Where all the variables are listed below (Figure 1 & Equation 1-4):

- $FP_t^{real}$: Ground truth full-view projection
- $P_t^{syn}$: noisy synthesized projection at time $t$
- $P_0^{syn}$: clean synthesized projection (final output of reverse process)
- $P_0^{real}$: Limited-angle projections ($135°: 225°$, conditional input)
- $\overline{\alpha_t}$: cumulative noise factor, $\overline{\alpha_t} = \prod_{s=1}^{t} \alpha_s$
- $\alpha_t$: noise schedule scalar at timestep $t$
- $\epsilon \sim \mathcal{N}(0, I)$: Gaussian noise
- $\epsilon_\theta(\cdot)$: neural network prediction of noise
- $\sigma_t$: variance (or standard deviation) of reverse noise at step $t$
- $z \sim \mathcal{N}(0, I)$: Gaussian noise in reverse process
- $t \in [1, ..., T]$: timestep index in diffusion process
- $T$: total number of diffusion steps

### 2.1.3 Image-DDPM

The second DDPM module is Image-DDPM module (Image-DDPM shown in Figure 1. (c)), which is designed to refine intermediate CBCT volumes reconstructed from synthesized projections, yielding high-fidelity CBCT images with enhanced anatomical consistency. Let $I_t^{syn}$ denote the noisy volumetric CBCT at diffusion timestep $t$, and let $I_t^{det}$ be the deterministic intermediate CBCT volume reconstructed from the synthesized projections via GTM. The model is trained to denoise $I_t^{syn}$ in a reverse stochastic process conditioned on $I_t^{det}$, ultimately producing the final reconstruction $I_0^{syn}$. The forward diffusion process adds Gaussian noise to the ground truth CBCT image $I_0^{syn}$, producing a noisy sample $I_t^{syn}$ over $T$ steps:

$$q\left(I_t^{cycle} \middle| I_0^{real}\right) = \mathcal{N}\left(I_t^{cycle}; \sqrt{\overline{\alpha_t}} I_0^{real}, (1 - \overline{\alpha_t})I\right) \tag{5}$$

Where $\overline{\alpha_t} = \prod_{s=1}^{t} \alpha_s$ accumulates the noise schedule. In the reverse process, the denoising network $\epsilon_\theta$ learns to estimate the noise added in the forward process, conditioned on the intermediate volume $I_t^{det}$. The reverse sampling at each step is given by:

$$I_{t-1}^{syn} = \frac{1}{\sqrt{\alpha_t}}\left(I_t^{syn} - \frac{1 - \alpha_t}{\sqrt{1 - \overline{\alpha_t}}} \cdot \epsilon_\theta(I_t^{syn}, t, I_t^{det})\right) + \sigma_t \cdot z, \quad z \sim \mathcal{N}(0, I) \tag{6}$$

The model is trained using the standard simplified objective:

$$\mathcal{L}_{Recon-DDPM} = \mathbb{E}_{I_0^{real}, \epsilon, t}\left[\left\|\epsilon - \epsilon_\theta\left(I_t^{(cycle)}, t, I^{(rec)}\right)\right\|_2^2\right] \tag{7}$$

where $I_0^{real}$ is the ground-truth CBCT image used for supervision. This enables the model to learn to reconstruct a clean anatomical image from corrupted volumes, guided by structural cues from GTM. Where all the variables are listed below:

- $I_0^{real}$: Ground-truth CT image reconstructed from full-view projections
- $I_t^{syn}$: Noisy CT volume at diffusion step $t$ (during reverse process)
- $I_0^{syn}$: Final CT output synthesized by Image-DDPM
- $I_t^{cycle}$: Noisy CT volume at timestep $t$, reconstructed from forward-projected synthetic projection via GTM during cycle supervision
- $I_0^{cycle}$: Final output of Image-DDPM, representing the sCT
- $I^{rec}$: Intermediate reconstruction from GTM, used as strict condition
- $\epsilon \sim \mathcal{N}(0, I)$: Gaussian noise
- $\epsilon_\theta \left( I_t^{(cycle)}, t, I^{(rec)} \right)$: Predicted noise in volume domain by Image-DDPM network
- $\mathcal{L}_{Recon-DDPM}$: Loss function used to train the Image-DDPM.

### 2.1.4 Geometry Transformation Module (GTM)

The fixed, differentiable Geometry Transfer Module (GTM) links projection and image spaces for the entire pipeline. In the forward direction, it uses a pixel-driven projector to convert a CBCT volume into simulated X-ray views; these are then passed back through GTM's inverse step to reconstruct a CBCT volume that enforces cycle consistency. In the inverse direction, GTM performs filtered back-projection on full-view projections, first producing an intermediate volume that conditions Image-DDPM and later recovering a CBCT volume from synthetic projections for the cycle-domain loss. We adopted a LEAP-CT FDK² with a ramp filter cutoff frequency of 1.0 in PyTorch so the system remains fully differentiable. We matched all projection and reconstruction parameters—such as source-to-isocenter distance, detector spacing, and view angles—to the geometry of the gantry-mounted CBCT system used in Siemens radiotherapy linacs. The following parameters were used in our GTM:

- Source-to-Detector Distance (SDD): 1500 mm
- Source-to-Isocenter Distance (SID): 1000 mm
- X-ray Detector Size: 768 × 1024
- X-ray Detector Spacing: 0.78 mm
- Projection Angles: 0°–360°
- Reconstruction Diameter: 495 mm
- Pixel Spacing: 0.9668 mm (in-plane)
- Slice Thickness: 1.0 mm
- Matrix Size: 512 × 512

These geometry parameters are applied consistently across projection synthesis and CBCT volume recovery, ensuring that simulated views adhere to real-world scanner configurations. Although GTM itself is not trainable, its integration guarantees that the LA-GICD framework remains physically grounded and spatially consistent across all modules.

## 2.1.5 GICD Strategy: Geometry-Integrated Cycle-Domain Supervision

The GICD supervision strategy integrates physical acquisition geometry and bidirectional consistency to guide training under limited-angle settings. It consists of three core components: Geometry-Conditioned Forward Path. Geometry-Conditioned reconstruction, the Projection-DDPM generates full-view synthetic projections $P_0^{\text{syn}}$ from limited-angle real projections $P_0^{\text{real}}$. These are passed through the GTM to reconstruct intermediate CBCT volumes $I^{\text{rec}}$, which serve as structural conditioning for the Image-DDPM. To ensure that $I^{\text{rec}}$ is physically plausible and aligns with the actual CBCT geometry, it is supervised by a ground truth volume $I^{\text{real-rec}}$ reconstructed from the real full-angle projections via the same GTM. The loss is defined as:

$$\mathcal{L}_{CT-rec} = MAE(I^{real-rec}, I^{rec}) \tag{8}$$

The final synthetic CBCT output $I_0^{\text{cycle}}$ generated by the Image-DDPM is directly supervised against the ground truth full-scan CT volume $I_0^{\text{real}}$. This supervision is computed using a voxel-wise mean absolute error:

$$\mathcal{L}_{CT-cycle} = MAE\left(I_0^{real}, I_0^{cycle}\right) \tag{9}$$

This path ensures that the entire pipeline, from limited-angle projection synthesis to final CBCT volume generation, is grounded in anatomical reality. The total loss function from GICD used during training is:

$$\mathcal{L}_{CT} = \mathcal{L}_{\theta I}^{\mu} + \gamma_1 \mathcal{L}_{\theta I}^{\Sigma} + \gamma_2 \mathcal{L}_{CT-rec} + \gamma_3 \mathcal{L}_{CT-cycle} \tag{10}$$

Where $\mathcal{L}_{\theta I}^{\mu}$ and $\mathcal{L}_{\theta I}^{\Sigma}$ denote the mean and variance prediction losses in Image-DDPM, and the weights are empirically set as $\gamma_1 = 0.05$, $\gamma_2 = 0.5$ and $\gamma_3 = 0.5$.

### 2.1.6 Inference Procedure

At inference time, as shown in Figure 1, our pipeline operates in a cascaded fashion, beginning with limited-angle projections and culminating in a fully synthesized CBCT volume. First, the Projection-DDPM module takes the limited-angle projection $P_0^{\text{real}}$ as conditioning input and generates a full-view projection $P_0^{\text{syn}}$ via the reverse diffusion process, starting from pure Gaussian noise. This synthesized projection is then passed through the GTM, which applies the imaging geometry extracted from DICOM metadata to reconstruct an intermediate CBCT volume $I^{\text{rec}}$. Next, the Image-DDPM module takes $I^{\text{rec}}$ as its conditioning input and refines it through a second diffusion process. Starting again from Gaussian noise, the model iteratively denoises the sample using the learned volume-domain prior, ultimately yielding the final synthetic CBCT volume $I_0^{\text{cycle}}$. This inference procedure eliminates the need for full-view measurements at test time and enables reconstruction from limited-angle inputs alone.

### 2.1.7. Implementation Detail

All models were implemented using PyTorch and trained on a single NVIDIA A100 GPU with 80 GB memory. Both Projection-DDPM and Image-DDPM were trained using 1000 diffusion steps during learning and sampled using 50 inference steps at test time. A cosine-based noise schedule $[\beta_t]_{t=1}^{T}$ was used, with cumulative noise factor $\overline{\alpha_t} = \prod_{s=1}^{t} \alpha_s$. The reverse diffusion followed a denoising diffusion implicit models (DDIM) style reparameterization using predicted noise and learned variances. During training, a batch size of 1 was used due to the high memory demand of 3D volume modeling. The optimizer was

AdamW with a learning rate of $2 \times 10^{-5}$ and default $(\beta_1, \beta_2) = (0.9, 0.999)$, and no weight decay. All experiments used fixed-angle limited projection inputs acquired over a 90° arc spanning 135° to 225°. The Projection-DDPM was trained to synthesize full 360° projections from these limited-view measurements. Input intensities (for both projection and CBCT volume domains) were linearly rescaled to the normalized interval [−1,1]. This mapping was applied consistently across training and inference. Spatial resolution was kept at 512×512, with pixel spacing retained from the original DICOM. Preprocessing included DICOM metadata parsing to extract geometric parameters such as source-to-detector distance (1500 mm), source-to-patient distance (1000 mm), and detector size, all of which were integrated into the GTM for accurate forward and backward projection. During inference, the Projection-DDPM synthesizes full-view projections conditioned on limited-angle inputs. These projections are passed through GTM to generate intermediate reconstructions, which are further refined by the Image-DDPM using geometry-informed cycle-domain priors. Both DDPM modules follow a U-Net backbone, with domain-specific conditioning.

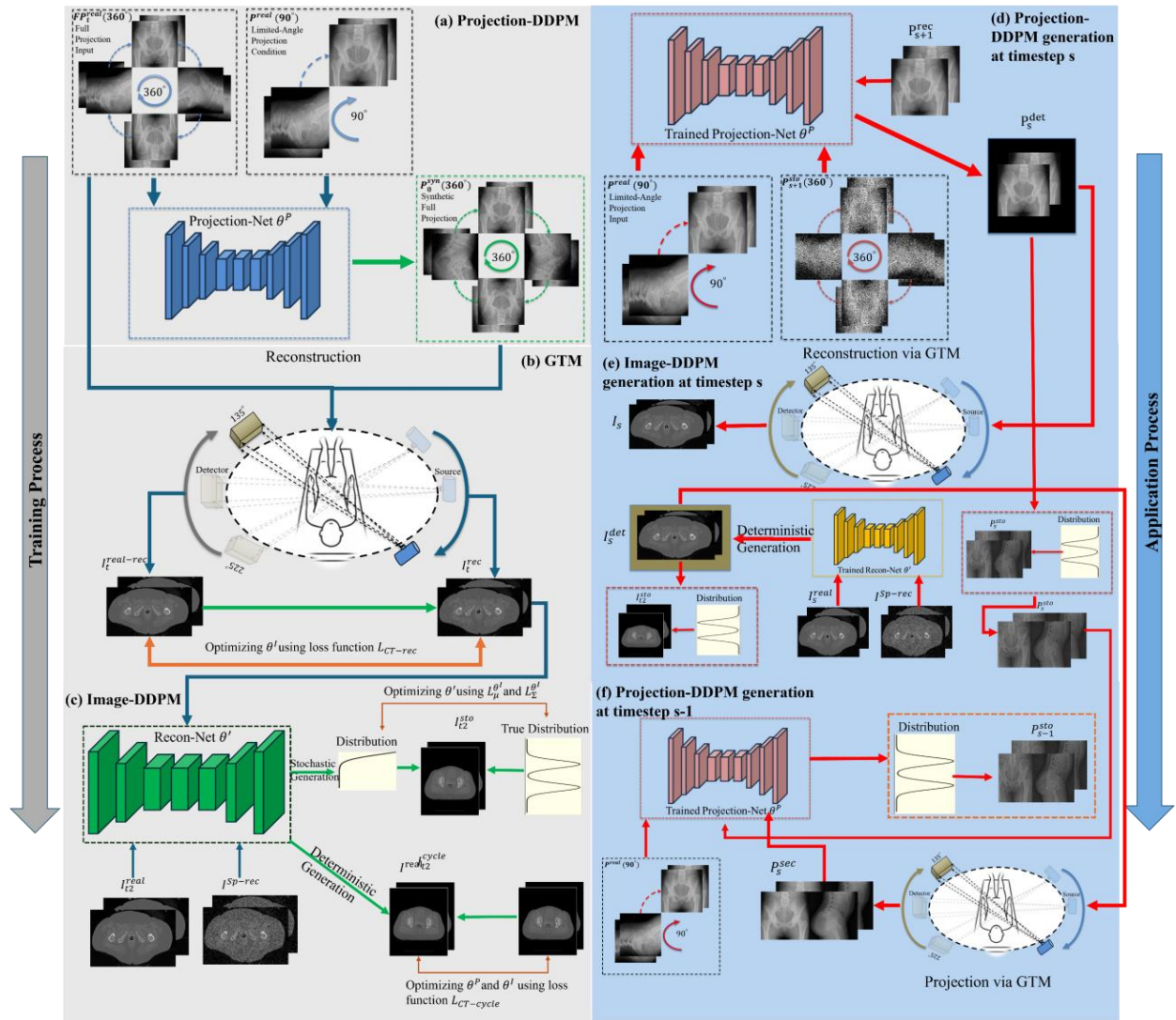

**Figure 1**. Overview of the training and application pipeline of the proposed dual-stage diffusion-based limited-angle CBCT reconstruction framework. (a) Projection-DDPM training, the first stage learns to synthesize full 360° projection stacks from limited-angle inputs (from 135° to 225°, total 90° projections). Given the paired real projections $P_t^{real}$ (360° projections) and limited-angle input $P^{real}$ (90° projections), the Projection-Net $\theta^P$ is trained to produce a pseudo full-angle projection

$P_t^{syn}$ (360° synthetic projections), which serves as a geometric prior for subsequent reconstruction. (b) GTM-based reconstruction supervision. The GTM module simulates the forward projection process based on detector–source geometry. By feeding both $P_t^{real}$ and $P_t^{syn}$ into GTM, corresponding reconstructed CBCT images $I_t^{real-rec}$ and $I_t^{syn-rec}$ are obtained. (c) Image-DDPM training, the second-stage generative module, Image-DDPM, learns to refine coarse CT volumes into anatomically faithful reconstructions. It receives GTM-reconstructed images from synthetic projections ($I^{sp-rec}$) as conditional inputs. During training, Image-DDPM performs reverse denoising to model the conditional posterior $P_\theta(x_0|I^{sp-rec})$, guided by structural consistency with the ground-truth CBCT $I^{ref}$. A dual-path training strategy is employed: the deterministic path supervises predictions from $I^{real-rec}$, while the stochastic path enables generation from $I^{sp-rec}$. In addition to the diffusion loss $\mathcal{L}_{DDPM}$, a cycle-consistency loss $\mathcal{L}_{CT-cycle}$ is applied to align reconstructions from synthetic projections with those from real ones. (d) Projection-DDPM generation at timestep $s$, during inference, a limited-angle projection set $P_{90°}^{real}$ is provided as the conditioning input to the trained Projection-DDPM. Starting from Gaussian noise, the model performs reverse denoising to generate full-view pseudo projections. At an intermediate timestep $s$, the current projection estimates $P_s^{rec}$ is shown. This partial output represents a stochastic intermediate state of the diffusion trajectory toward the final 360° projection $\hat{P}_{0-360}$. (e) Image-DDPM generation at timestep $s$. Given a conditional input $I_t^{real-rec}$, which is reconstructed from limited-angle real projections via GTM, the trained Image-DDPM iteratively denoises a Gaussian latent to synthesize the target CBCT slice. At an intermediate timestep $s$, the model outputs a partially denoised estimate $I_s$. This prediction progressively approaches the final output $\hat{I}_t$ through the reverse diffusion trajectory. The training objective ensures that such samples remain structurally aligned with the ground truth $I_t$, enabling high-fidelity reconstruction under geometry-aware conditioning. (f) Projection-DDPM generation at timestep $s-1$. This diagram illustrates the reverse denoising process within Projection-DDPM at timestep $s-1$. Starting from Gaussian noise, the model iteratively predicts pseudo projections conditioned on the limited-angle input $P^{real}$. At each step, a sample $P_{s+1}^{sto}$ is transformed into the next state $P_s^{rec}$, approaching the final full-view distribution. These intermediate predictions can be optionally reconstructed into CBCT slices via GTM to assess geometric consistency during sampling. This stepwise evolution highlights the generative path that connects noise to structure-aware projections within the diffusion model.

## 2.2 Dataset and Evaluation Protocol

### 2.2.1 Dataset description

We studied 18 anonymized patients who received gynecologic HDR brachytherapy. The dataset comprises a total of 78 CT volumes, acquired from 18 patients who received 4 or 5 CBCT scans during the course of treatments. Each case provided a 512 × 512 × 256 pelvic CBCT volume with voxel spacing of 1 × 1 × 2 mm in native resolution; we kept the x–y spacing unchanged and linearly down-sampled the z-axis to 256 slices to fit GPU memory. All CT volumes were normalized to the [0,1] range, and the simulated projections were normalized the same way; at test time the reconstructed CBCTs were rescaled to their original Hounsfield-unit range for quantitative evaluation. We checked every volume to make sure the uterus, cervix, and adjacent soft tissue are intact. Using LEAP-CT[24], we built patient-specific cone-beam geometry from the DICOM tags and generated full-arc projections. For model input, we kept only 90 projection views that covered a 90° arc (135°–225°, clockwise), a range that reflects realistic limited-angle projections. The full 360° projections were generated for supervision, but we didn't expose to the model at inference. LEAP-CT then supplied the same geometric parameters—source-to-detector distance, detector pitch, slice thickness—to the GTM.

### 2.2.2 Data Splits and Evaluation Metrics

To evaluate the generalizability of the proposed model across anatomical variations and treatment fractions, a volume-level split was employed rather than a patient-level separation. Specifically, a total of 78 volumes were collected from 18 patients. To ensure patient-level independence in model evaluation, 60 volumes were randomly chosen for training, and one representative volume per patient (18 volumes) was held out for testing. No patient overlap exists between the training and testing subsets to prevent data leakage.

The reconstructed CBCT volumes were quantitatively evaluated using standard image similarity metrics, including:

- Peak Signal-to-Noise Ratio (PSNR), expressed in decibels (dB), to measure pixel-level fidelity.
- Structural Similarity Index Measure (SSIM), computed over 11×11 windows to assess structural preservation.
- Mean Absolute Error (MAE), calculated voxel-wise over the body region to evaluate global intensity error.

For each metric, values were computed between the model output (Limited-angle reconstructed CBCT) and the corresponding ground truth CBCT volume. Unless otherwise noted, all metrics were reported as averages over the full test set.

### 3. Quantitative and qualitative evaluation

We evaluated LA-GICD on a limited-angle CBCT dataset using three standard metrics: MAE, SSIM, and PSNR. As shown in Table 1, the proposed method demonstrates strong reconstruction performance under severe angular limitations. These results are averaged across all test cases and anatomical views, including axial, coronal, and sagittal planes.

**Table 1**. Quantitative comparison of LA CBCT reconstructions using LA-GICD and conventional FDK. MAE, SSIM, PSNR were evaluated on 18 test volumes. LA-GICD significantly outperformed FDK in all metrics ($p < 0.01$, paired $t$-test), demonstrating superior image fidelity and structural consistency under 90° limited-angle acquisition.

|  | MAE (HU) | SSIM | PSNR (db) |
|---|---|---|---|
| **LA-GICD** | 35.51±5.324 | 0.84±0.02 | 29.79±0.95 |
| **FDK** | 180.70 ± 33.85 | 0.54 ± 0.04 | 19.91 ± 1.24 |
| **Paired t-test (LA-GICD vs FDK)** | <0.01 | <0.01 | <0.01 |

Visual comparisons under axial, coronal, and sagittal views are presented in Figures 2–4, showcasing results from eight representative test cases. Each column corresponds to a different patient. For each case, we display the reconstruction from full-angle data (ground truth), the proposed LA-GICD, and the direct reconstruction from limited-angle projections, followed by the corresponding error maps. LA-GICD visibly reduces shading and streak artifacts and soft-tissue blurring across all views. Compared to FDK baseline, our method better preserves organ boundaries and low-contrast structures, with lower error magnitudes and more uniform residual maps.

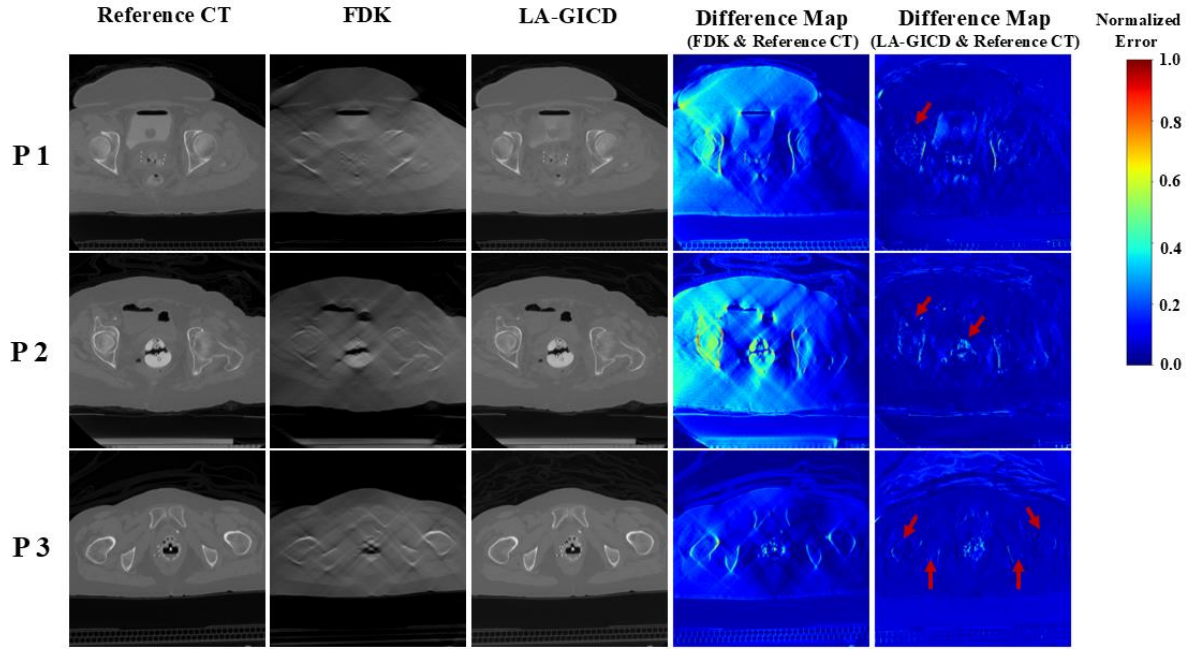

**Figure 2**. Axial slice comparison of CBCT reconstruction quality using LA-GICD versus conventional FDK under limited-angle acquisition over three patients. Each row corresponds to a different patient. From left to right: ground truth (360° CBCT), limited-angle FDK, LA-GICD reconstruction from limited-angle projections, voxel-wise error maps, and hallucination-prone residual maps highlighting local discrepancies. Compared to FDK, LA-GICD markedly reduces streaking artifacts and recovers soft-tissue and bony structures with improved fidelity.

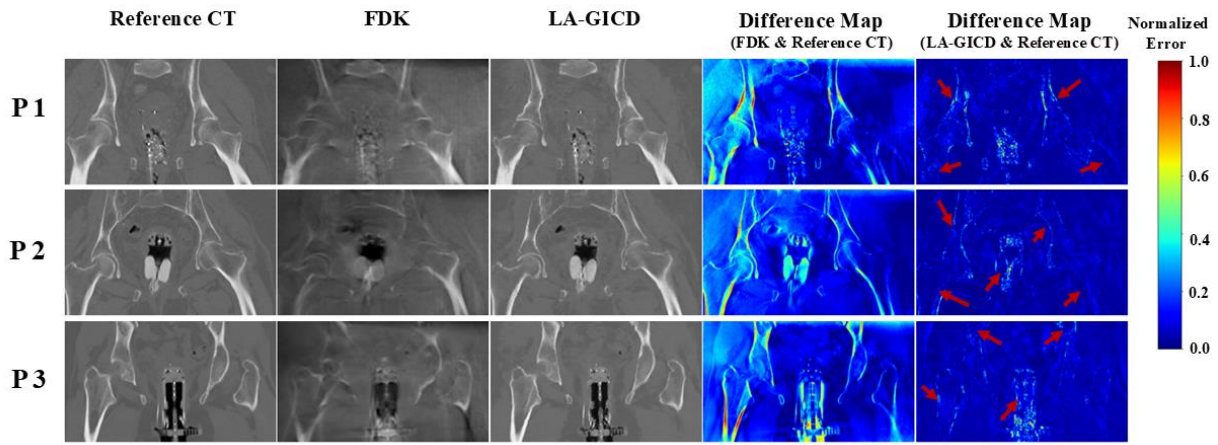

**Figure 3**. Coronal slice comparison of CBCT reconstruction quality using LA-GICD versus conventional FDK under limited-angle acquisition over three patients. Each row corresponds to a different patient. From left to right: ground truth (360° CBCT), limited-angle FDK, LA-GICD reconstruction from limited-angle projections, voxel-wise error maps, and hallucination-prone residual maps highlighting local discrepancies. Compared to FDK, LA-GICD markedly reduces streaking artifacts and recovers soft-tissue and bony structures with improved fidelity.

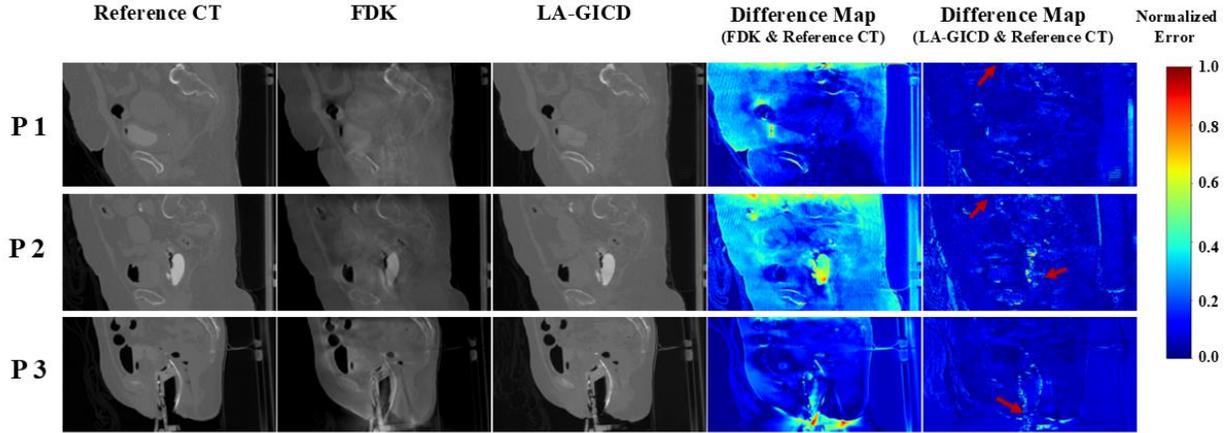

**Figure 4**. Sagittal slice comparison of CBCT reconstruction quality using LA-GICD versus conventional FDK under limited-angle acquisition over three patients. Each row corresponds to a different patient. From left to right: ground truth (360° CBCT), limited-angle FDK, LA-GICD reconstruction from limited-angle projections, voxel-wise error maps, and hallucination-prone residual maps highlighting local discrepancies. Compared to FDK, LA-GICD markedly reduces streaking artifacts and recovers soft-tissue and bony structures with improved fidelity.

## 4. Discussion

Limited-angle CBCT has the potential to significantly improve clinical workflows by reducing scan time, minimizing motion artifacts, and avoiding physical gantry collisions—particularly in RT settings such as equipment packed HDR rooms, intra-fractional motion managements. However, reconstructing high-quality volumes from such limited data remains a major challenge. This study demonstrates that a geometry-integrated, dual-domain diffusion framework can reconstruct high-quality CBCT volumes from a single 90° arc of projections. With one model trained once on this limited-angle configuration, LA-GICD achieved a mean absolute error of 35 HU, an SSIM of 0.84, and a PSNR of 29.8 dB on 18 gynecologic HDR brachytherapy scans. As illustrated in Figures 2–4, the method markedly reduced shading/streak artifacts and preserved soft-tissue detail across axial, coronal, and sagittal views, despite the combined challenges of limited angle coverage and metallic applicators. The performance gain stems from the GICD training scheme, where a Projection-DDPM and an Image-DDPM are optimized jointly with a single optimizer. In each training step the Projection-DDPM infers the missing views, a fixed analytic cone-beam back-projector converts the completed sinogram into an image, the Image-DDPM denoises and sharpens this volume, and the refined image is forward-projected back to projection space. The resulting projection-image-projection loop is differentiated end-to-end, anchoring every update to the measured data while steadily suppressing hallucinated structures.

Early deep learning approaches treated limited-angle reconstruction as a post-processing task[13,14], where a CNN or U-Net is trained to map artifact-contaminated images reconstructed from limited-angle projections to high-quality CT images obtained from full-angle scans. Because raw projections are never revisited, post-processing CNNs cannot enforce projection domain consistency. Residual streaks often remain along the direction of missing angles, and the learned priors may generate anatomically realistic but incorrect structures, especially when the angular coverage changes or high-density hardware is present. In contrast, our framework operates in both the projection and image domains. The Projection DDPM completes the projections, the Recon DDPM refines the image, and a fixed cone-beam projector and back-projector connect the two. This dual-domain geometry-informed approach reduces artifacts and lowers the risk of hallucinated anatomy. To improve reconstructed image fidelity, later studies unfolded model-based iterative reconstruction into a fixed sequence of trainable stages. Each stage applies a data-consistency update followed by a learned regularization step, and all parameters are optimized end-to-end from paired full-angle and limited-angle data[15,16]. Although this design suppresses artifacts more effectively than image-only CNNs (post-processing approach), its performance is still highly sensitive to the choice of step size and regularization weights and often depends on training datasets that match the specific scan geometry and angular deficit. When the acquisition conditions differ from those used during training, residual streaks and staircase effects frequently reappear. These limitations reveal the need for a reconstruction scheme that generalizes across angular coverage without retraining. Because LA-GICD decouples projection completion and image refinement from any fixed view range, a single network could in principle be trained on mixed-angle data and then applied to previously unseen arc lengths—a possibility we plan to evaluate in future work. Most recently, score-based DDPM has been explored for sparse and limited-angle CT reconstruction[19,25]. These models learn the gradient (score) of the data distribution and then iteratively transform pure Gaussian noise into a synthetic image or projection that matches the training data. Because the denoising is applied at multiple noise scales, diffusion models can recover high-frequency details that earlier networks often blur. Their main limitation in limited-angle CT is insufficient geometric conditioning. In most implementations the model is conditioned only on the incomplete projections (or on a coarse FBP/FDK image derived from it); the forward–projection operator and exact ray geometry are not embedded in the reverse process. As a result, the diffusion trajectory has no explicit penalty for deviating from the true measurement space inside the missing wedge. During inference the model can therefore inpaint this region with structurally reasonable but physically inconsistent content, especially when the acquisition arc, source–detector distance, or patient positioning differ from those represented during training. LA-GICD is explicitly designed to overcome the geometric-conditioning gap identified above. First, the framework embeds a fixed analytic cone-beam projector and back-projector (GTM) within the diffusion loop, so every denoising step is evaluated against the true ray geometry rather than an implicit image prior. Second, the projection-image-projection cycle forces the Projection-DDPM and Image-DDPM to agree with the measured data at each iteration, sharply limiting the degrees of freedom available for hallucination. Third, because the two DDPMs are optimized jointly under the same loss, projection completion and image refinement co-evolve, allowing geometric errors in one domain to be corrected by feedback from the other. In practice, these features yield streak-free reconstructions that remain anatomically accurate even in the presence of metal applicators and across different patient anatomies, while maintaining the fine detail recovery characteristic of diffusion models. Taken together, these design elements enable LA-GICD to combine the anatomical sharpness and generative flexibility of diffusion models with the geometric rigor of traditional forward–inverse operators. Unlike prior approaches that rely solely on image priors or approximate data consistency, LA-GICD enforces projection fidelity at every step, explicitly resolves missing-angle ambiguity, and preserves structural realism even under challenging clinical conditions. This dual-domain, geometry-integrated strategy offers a promising pathway toward generalizable limited-angle CT reconstruction that is robust to scan variability, patient heterogeneity, and metallic implants—without retraining or view-specific tuning.

Clinically, the proposed LA-GICD framework addresses several longstanding barriers in image-guided brachytherapy and other interventional workflows. From a procedural standpoint, in-room CBCT provides immediate volumetric feedback at the point of care, eliminating the need to transfer the patient between imaging and treatment rooms. In HDR gynecologic treatments, patients are frequently positioned in lithotomy or oblique orientations that severely restrict gantry clearance. Under these conditions, a full CBCT scan is often infeasible due to the risk of collisions with the couch, patient, or other equipment. By limiting acquisition to a single 90° arc, LA-GICD avoids mechanical interference while preserving volumetric imaging capability. This is particularly beneficial in cases involving metallic applicators or extended transperineally needles, where conventional CBCT is either skipped entirely or replaced by off-table imaging. This enables the clinical team to verify applicator geometry, evaluate target coverage, and make intra-procedural adjustments without disrupting the sterile field or prolonging anesthesia time. The shorter scan duration—approximately one-quarter of a conventional full-arc CBCT—also reduces the likelihood of motion-induced artifacts caused by respiration, bowel peristalsis, or patient discomfort during external beam radiotherapy. As a result, LA-GICD facilitates the acquisition of high-fidelity online images with minimal motion blur, even in anatomically unstable regions such as the pelvis and abdomen. Moreover, the substantial reduction in imaging dose—without loss of image quality—supports repeated on-board imaging throughout the treatment course, particularly relevant in pediatric patients or those requiring multiple fractions within a short treatment window. Because the LA-GICD algorithm operates as a post-acquisition reconstruction module, compatible with existing flat-panel CBCT systems, it introduces no additional hardware requirements. This software-only implementation facilitates broad deployment across diverse clinical settings, including high-throughput centers and resource-limited environments, potentially expanding access to adaptive and image-guided brachytherapy.

Hallucination—generation of anatomically plausible yet physically incorrect structures—is an inherent concern when using deep generative models such as DDPM for limited-angle CBCT reconstruction. To systematically evaluate this risk, we conducted repeated reconstructions from identical projection data sets and examined voxel-wise uncertainty (Fig. 5). Uncertainty maps showed slightly elevated variability adjacent to steep attenuation gradients, such as at bone–soft-tissue interfaces and around metallic applicators, which are prone to reconstruction errors. However, the magnitude of these deviations remained low (typically less than 0.01 normalized units), indicating that LA-GICD preserves high fidelity and effectively suppresses hallucinations even in the most challenging regions. Nevertheless, acknowledging the presence of these minor inconsistencies highlights the necessity of incorporating uncertainty quantification into clinical decision-making. Future studies could further reduce such uncertainty by embedding additional physics-based constraints, explicit anatomical priors, or uncertainty-aware training strategies into the diffusion framework.

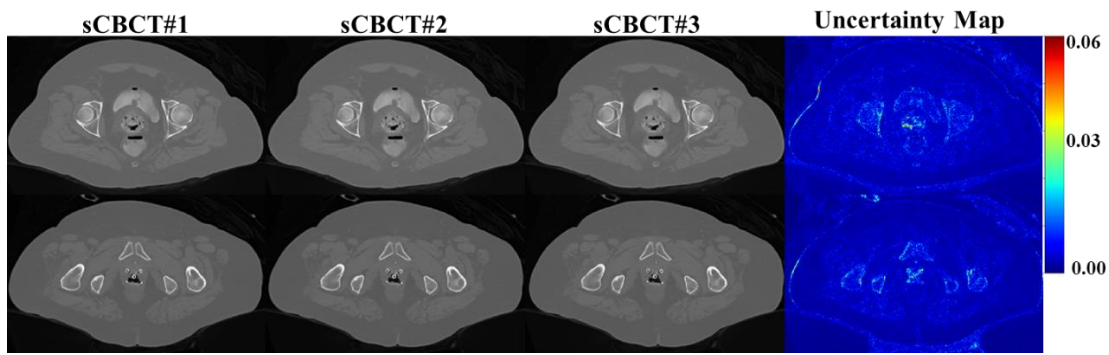

**Figure 5**. Demonstration of hallucination variability in LA-GICD limited-angle CBCT reconstruction. Three independent reconstructions (columns 1–3) of the same patient using LA-GACD with identical limited-angle inputs reveal visible inconsistencies in soft-tissue regions, particularly near applicator tips and bladder boundaries. The fourth column visualizes the normalized corresponding voxel-wise standard deviation across runs, highlighting high-uncertainty regions (0–0.06 range) in a jet colormap.

This study has several limitations. First, our validation focused solely on limited-angle CBCT reconstruction for gynecologic HDR brachytherapy. Although these data provided strong evidence of clinical utility for pelvic anatomy, the performance of our proposed LA-GICD framework in other anatomical regions—such as thoracic, abdominal, or head-and-neck regions—remains unknown. Different anatomical sites vary significantly in complexity, internal motion characteristics, and tissue contrast; thus, the results observed in the pelvis may not directly extend to other clinical applications or anatomical contexts. Second, computational complexity remains a concern. Diffusion-based models, including our LA-GICD framework, inherently require iterative sampling procedures and multiple denoising steps during inference. Consequently, inference times are currently longer compared to standard analytical reconstruction methods (e.g., FDK) or simple convolutional neural network approaches. The prolonged reconstruction times might limit the practical deployment of the method, especially in real-time or intraoperative imaging scenarios, where timely feedback is crucial for procedural decision-making. Third, although our model demonstrated robustness to the specific 90° limited-angle configuration tested, we did not systematically evaluate the stability of image quality under different angular spans or orientations. Technically, the precise acquisition angle may vary due to patient positioning constraints, procedural setup, or equipment restrictions. Without explicit verification across multiple angular configurations (e.g., smaller arcs or different orientations), it remains unclear whether reconstruction accuracy and artifact suppression performance would remain consistent. Finally, the LA-GICD model offers the possibility of patient-specific fine-tuning, potentially enhancing reconstruction accuracy by leveraging patient-specific data from previous imaging fractions. However, such personalization requires collecting additional imaging data and computationally intensive retraining procedures for each individual patient. These requirements introduce practical barriers, potentially limiting widespread adoption in clinical workflows where computational resources, staffing, or imaging time are constrained.

Future research should evaluate the generalizability of LA-GICD to other anatomical sites and clinical settings beyond pelvic HDR brachytherapy. This includes thoracic and abdominal sites where internal organ motion, tissue heterogeneity, and respiratory artifacts may pose different challenges for limited-angle reconstruction. Expanding the training dataset to encompass diverse anatomical contexts could improve robustness and clinical utility. Further optimization is also needed to reduce inference time. Techniques such as model pruning, knowledge distillation, and cascaded denoisers may help accelerate the diffusion process without compromising image quality, enabling near real-time reconstruction suitable for time-sensitive workflows. To ensure reliability in variable acquisition settings, future studies should systematically test LA-GICD under a range of angular spans and orientations. Evaluating model stability across different arc lengths (e.g., 60° or 45°) and gantry paths would clarify their robustness in flexible clinical deployments. In addition, while LA-GICD demonstrated minimal hallucination in our internal assessments, rigorous prospective evaluation is warranted. This includes both qualitative review by expert clinicians and quantitative testing with anatomically altered phantoms or surgical validation data to confirm that high-frequency details are faithfully reconstructed from incomplete data. Finally, we aim to explore strategies for integrating uncertainty estimation into the reconstruction pipeline, allowing clinicians to identify image regions with low reconstruction confidence—an important safeguard when using AI-generated volumes in interventional and high-dose workflows.

## 5. Conclusion

We proposed a new LA-GICD framework for LA CBCT reconstruction. The method reconstructs high-quality CBCT volumes from a 90° short-arc projection data, without retraining across angles or patients. Tested on gynecologic HDR datasets, LA-GICD achieved strong quantitative accuracy and visibly reduced artifacts. This approach offers a practical limited-angle reconstruction solution for CBCT imaging in time-sensitive or geometrically constrained radiotherapy workflows.

### Acknowledgements

This research is supported in part by the National Institutes of Health under Award Number R01CA272991, R01EB032680 and U54CA274513.